# Denis Saklakov

Robotech Frontier Hub, New York

# Phase-Coded Memory and Morphological Resonance: A Next-Generation RAG Architecture

## Abstract

In this white paper, we introduce a cognitive Retrieval-Augmented Generator (RAG) architecture that transcends the context length limitations of transformer-based models by leveraging phase-coded memory and morphological-semantic resonance. The proposed system replaces token-based embeddings with frequency-phase representations of meaning, enabling knowledge to be stored as dynamic wave patterns. A three-tiered design is presented – a *Morphological Mapper* that encodes inputs into complex semantic waveforms, a *Field Memory Layer* that superposes and stores these waveforms as distributed holographic traces, and a *Non-Contextual Generator* that produces coherent outputs by tapping resonant semantic fields rather than sequential token contexts. We detail how phase interference and semantic resonance in the Field Memory allow direct content-addressable retrieval without conventional document ranking. The generator draws on resonant memory signals in real-time, freeing it from fixed context windows and linear order constraints. We discuss theoretical underpinnings and provide pseudocode-style illustrations of the mechanism. Finally, we highlight significant gains in efficiency – memory, energy, and time – achieved by eliminating massive context re-encoding and sequence scanning. Early experiments in phase-coded representations (e.g. holographic reduced representations, complex-valued RNNs, and wave-based semantic memory) are cited to demonstrate the feasibility and promise of this approach[1][2].

## Introduction

Transformer-based RAG systems have achieved impressive results in augmenting language models with external knowledge, yet they are fundamentally limited by context window size and token-based processing overhead[3][4]. Even state-of-the-art models are typically constrained to a few thousand tokens of context, meaning they cannot directly consider knowledge beyond this limit without resorting to fragmenting input or compressing information[3][5]. This limitation forces current RAG implementations to retrieve a small subset of documents or passages, rank them by similarity, and then concatenate them into the model's input context. The result is not only a hard cap on how much knowledge can be leveraged at once, but also significant computational redundancy – the model must repeatedly re-process lengthy text contexts for each query, and attention costs scale quadratically with input length[3][6]. Furthermore, representing knowledge as sequences of tokens (or static vectors derived from tokens) can be semantically brittle: subtle variations like negation or nuance are hard to capture with magnitude-based similarity alone[7][8]. For instance, in a conventional embedding space, the vectors for "happy" and "*not* happy" may end up quite close despite opposite meanings, highlighting how context-dependent semantics (like negation) are not explicitly encoded[8].

To overcome these challenges, we propose a phase-coded retrieval-augmented generator that moves beyond token contexts. Instead of encoding knowledge as discrete text or high-dimensional static vectors, our system encodes and stores information as wave patterns – functions with an amplitude and phase at each semantic dimension[1]. By operating in a wave domain, the system can exploit interference and resonance properties to retrieve relevant information, analogous to how holograms or neural oscillations retrieve content by pattern matching rather than brute-force search[9][10]. Crucially, generation is performed without inserting retrieved text into a fixed context window. The generator component accesses memory *on the fly* through resonance signals, producing outputs that remain coherent and grounded in the query's relevant semantic field, yet are not limited by any predetermined context length. This architecture aims to save memory and computation (since large contexts need not be repeatedly loaded or attended over) and to yield more robust semantic understanding (since phase-coded representations naturally encode structural relations like negation or emphasis as phase shifts).

In the following sections, we present the formal architecture of our system and explain each of its three tiers. We detail how meaning is encoded as frequency-phase patterns (bypassing traditional token embeddings), how the Field Memory Layer uses distributed holographic storage to superpose many patterns and retrieve them via interference, and how the Non-Contextual Generator composes an answer by continuously resonating with the memory field rather than reading a static context. We then discuss the advantages of this design in terms of efficiency and scaling. Finally, we connect our approach to prior research and models – from Holographic Reduced Representations in cognitive science to modern complex neural networks – that provide evidence for the feasibility of phase-based semantic memory and retrieval.

## Architecture Overview

Our proposed architecture is organized into a three-tier system that reflects the flow from input query to stored knowledge to output generation:

- Morphological Mapper: Transforms incoming language (queries or prompts) into a *morphological-semantic waveform* representation, encoding meaning using frequency and phase rather than discrete tokens.
- Field Memory Layer: A distributed memory store that holds semantic content as complex wave patterns in a shared field. It supports storage via superposition of waveforms and retrieval via phase-based interference (resonance) without explicit search ranking.

- Non-Contextual Generator: A generative model that produces the final answer by interacting with the Field Memory through resonance. It does not rely on a fixed token context; instead, it uses real-time semantic feedback from memory (the resonant patterns) to ensure coherence and factuality.

This design can be visualized as an encoder-field-decoder pipeline, but unlike traditional encoder-decoder models, the "encoder" here (Morphological Mapper) outputs a *pattern* that directly probes a *standing wave* memory, and the "decoder" (Generator) continually conditions on memory *signals* rather than a one-time encoded vector or sequence. We next describe each component in depth.

## Morphological Mapper (Phase-Based Encoder)

The Morphological Mapper is responsible for converting textual inputs into the phase-coded semantic representation used by the rest of the system. It replaces the standard token embedding and positional encoding steps of a transformer with a process that yields a complex valued pattern $\psi(x)$ for each input, where $\psi(x) = A(x) e^{i\phi(x)}$[1]. Here $A(x)$ is an amplitude function capturing the *strength or salience* of semantic features at dimension $x$, and $\phi(x)$ is a phase function encoding the *contextual or relational aspect* of those features[11][8]. In simpler terms, amplitude corresponds to "how strongly this concept is present," while phase corresponds to "in what semantic mode or context this concept appears." Crucially, this approach eschews token-based embeddings – we do not represent a word by a single point in a vector space. Instead, each meaningful unit (word, morpheme, or concept) is mapped to a wave pattern distributed across dimensions.

Morphological analysis is integral to this component. Linguistic morphology (prefixes, suffixes, roots, negations, etc.) is leveraged to modulate the phase of the base semantic representation. For example, consider the words "happy" vs. "unhappy." A traditional encoder might produce two embeddings that end up near each other in vector space. In our system, the mapper would recognize the negating morpheme "un-" and adjust the phase of the "happy" concept wave by $\pi$ radians (180°) to encode the opposition in meaning[8]. The resulting $\psi_{\text{happy}}(x)$ and $\psi_{\text{unhappy}}(x)$ might share the same amplitude profile $A(x)$ (indicating the same base concept), but their phase functions $\phi(x)$ differ by $\pi$ in appropriate subspaces, making them out of phase in the semantic field. When these patterns later interact in memory, their phase difference will cause destructive interference (canceling out in overlapping dimensions) signifying semantic negation[12][13]. This illustrates how morphological and syntactic cues (negation, modality, tense, etc.) are converted into phase patterns that modify meaning representation in a structured way.

Formally, the Morphological Mapper can be seen as implementing a function: $$\text{Mapper}(T) \rightarrow \psi,$$ where $T$ is a token sequence (the query or prompt) and $\psi$ is the complex wave pattern representing its meaning. The mapping involves several steps: 1. Linguistic Parsing: Identify meaningful units such as roots, affixes, and compounds in the input text. For instance, split "nationality" into "nation + -ality" or detect that "did not go" carries negation on "go". 2. Base Vector Assignment: For each identified base concept or morpheme, retrieve an initial high-dimensional vector $v \in \mathbb{R}^n$ (as one would in a normal embedding space). These could be learned embeddings from a prior model or one-hot encodings in a semantic vector space. 3. Phase Encoding: Transform each base vector into a complex pattern. A simple scheme (compatible with existing embeddings) is: set amplitude $A(x) = |v(x)|$ (the magnitude of each component) and phase $\phi(x) = 0$ if $v(x) \ge 0$ or $\pi$ if $v(x) < 0$[14]. This "sign-to-phase" mapping is one straightforward approach used in prior work to introduce phase while preserving compatibility with real-valued embeddings[14]. More sophisticated mappings could assign phase based on specific morphemes (e.g. a prefix "un-" might add $\pi$ to the phase of certain semantic features globally). 4. Waveform Composition: If multiple components (words/morphemes) are present, their wave patterns are superposed (added) to form a combined pattern for the whole input. However, simply adding can lead to interference even between parts of the query; to mitigate this, the Mapper can allocate separate frequency bands or phase offsets for distinct sub-components (analogous to multiplexing signals) so that the combined pattern retains distinguishable features for each constituent. This *morphological multiplexing* ensures that, for example, subject and object concepts in a query do not overwrite each other's information in the wave representation.

The output of the Morphological Mapper is a high-dimensional complex vector (or waveform) $\psi_{\text{query}}$ that encodes the entire query's semantics. Importantly, this representation does not have a sequence length or position in the traditional sense – it is more like a *distributed hologram* of the query's meaning. All information is smeared across the dimensions (like a frequency spectrum), enabling the next layer to treat the query as a probe signal that can be matched against stored memory patterns.

## Field Memory Layer (Holographic Wave Store)

The Field Memory Layer is a distributed semantic memory that stores knowledge as wave patterns and retrieves information via resonance. This layer draws inspiration from holographic memory in cognitive science and neuroscience – the idea that multiple memories can co-exist in a superposed form, and a cue can retrieve a specific memory by pattern interference rather than by index lookup[9][10].

Storage as Wave Interference: Each unit of knowledge (which could correspond to a document, a fact, an image's description, etc.) is encoded in the same manner as a query – as a complex pattern $\psi_{\text{memory}} = A_m(x)e^{i\phi_m(x)}$ – and stored in the Field Memory. Physically, we can imagine the Field Memory as a large array or "fabric" of complex numbers. Storing a pattern involves adding its complex vector to this fabric (much as holographic reduced representations store associations by superposition of convolved vectors[15][16]). If multiple memory items are stored, their patterns sum linearly in the field: $$\Psi_{\text{field}}(x) = \sum_{k} \psi_k(x),$$ where each $\psi_k$ is a stored item's wave pattern. This superposition principle is powerful – it means the memory can hold many items without allocating separate slots for each, effectively compressing information. The trade-off is that the patterns can interfere with each other, but if they are encoded properly (e.g. random phase distributions or orthogonalized patterns), the interference for non-matching retrievals appears as quasi-random "noise" that does not significantly affect the matching process[15][17]. Indeed, Plate (1995) showed that if vectors are high-dimensional and random, the superposition (he called it a *memory trace*) can store multiple associations such that retrieving with the correct cue yields the target item plus only minimal noise from others[16][15]. In our wave context, this implies that as long as stored patterns are sufficiently decorrelated in their phase alignment, they will not strongly interfere except with the correct query.

Retrieval by Resonance (Phase Interference): When a query's wave pattern $\psi_{\text{query}}$ (from the Morphological Mapper) is presented to the Field Memory, retrieval is accomplished not by iterating over items or computing discrete similarities, but by leveraging wave interference. The query pattern is injected into the field – mathematically, we can form the sum $\Psi_{\text{probe}}(x) = \psi_{\text{query}}(x) + \Psi_{\text{field}}(x)$ – and then observe the result of this interference. If $\psi_{\text{query}}$ closely *matches* one of the stored patterns (say $\psi_j$) in frequency content and phase alignment, it will constructively interfere with $\psi_j$ while largely destructively interfering or being orthogonal to other patterns[13][18]. Constructive interference means that at dimensions where phases align, amplitudes reinforce each other, producing a spike in the combined waveform's energy. Destructive interference means misaligned phases cancel out, diminishing the energy. The Field Memory can thus compute a resonance score or activation for each stored pattern (implicitly, in parallel) by measuring how much the combined wave's amplitude is amplified relative to baseline. The pattern with the highest resonance is effectively the retrieved item. In practice, one can derive a formula: if $S(\psi_a,\psi_b)$ is a similarity defined as half the normalized energy of $\psi_a + \psi_b$ (with a scaling factor to penalize energy mismatch)[19], then a query will retrieve item $j$ such that $j = \arg\max_k S(\psi_{\text{query}}, \psi_k)$, i.e. the memory pattern that produces maximum constructive interference with the query[19][20]. This $S$ function can be seen as a *phase-aware generalization* of cosine similarity[21]: it reduces to cosine when all vectors are real, but for complex waves it rewards phase alignment in addition to magnitude alignment[21].

Notably, this retrieval process does not require a separate ranking step over discrete documents – the physics of interference performs a type of content-based address. In a conventional vector store, one would compute similarity of the query vector to each stored vector and then rank the scores. Here, if the memory were implemented in analog hardware or a parallel architecture, one could in principle present the query as a signal to the memory and directly observe an echo or resonance from the matching memory location (much as a standing wave resonates at certain frequencies). The system *acts as if* all memories are queried simultaneously and only those with matching phase-frequency structure respond strongly. This is akin to how in a hologram, illuminating it with part of an image will cause the corresponding stored image to come into focus by interference, without scanning every stored image. Early cognitive models like Holographic Reduced

Representations captured a similar idea: using circular convolution as a binding operation, one can store multiple pairs of vectors in a single superposed vector and retrieve one component by convolving with the other[16][15]. The convolution/correlation operations in HRR essentially perform the resonance retrieval mathematically. Our Field Memory similarly encodes associations (between, say, a key and content) as phase relationships in a unified field, and uses interference (correlation in the complex domain) to retrieve the content given the key.

Encoding Semantic Content as Dynamic Waveforms: Each stored item's pattern $\psi_k(x)$ is not static – we can consider it a snapshot of a dynamic waveform that can oscillate or evolve. The Field Memory might be implemented as a recurrent dynamic system (e.g., a network of oscillatory elements) that naturally resonates at certain pattern frequencies. In fact, a connection can be made to associative memory in neural networks: traditional Hopfield networks store binary patterns as fixed-point attractors; by contrast, one can store oscillatory patterns (with phases) as *limit cycles* in a recurrent network[22][23]. Research by Biswas et al. (2021) demonstrated a complex-valued Hopfield-like network where each stored memory is a *stable oscillation* at a particular phase relationship across neurons[2][24]. Their network used multiple frequencies (Fourier-like decomposition of signals) and a special "power coupling" rule to achieve stable phase-locking for multiple coexistent oscillatory memories[2][24]. This provides a theoretical backing that a Field Memory could be realized as a neural circuit: each memory pattern is an oscillation (with a specific complex phase pattern) that can be reactivated or amplified if an input drives the network at that same phase pattern. In essence, the memory is like a field of oscillators tuned to different frequencies/phase-patterns; a query excites the field, and only those oscillators that match the query's frequency content resonate strongly (others remain at baseline or get dampened).

From a systems perspective, the Field Memory might also be implemented in software via fast mathematical operations. For instance, computing the resonance score $S(\psi_q,\psi_k)$ for all $k$ could be accelerated by FFT-based convolution if we interpret phase patterns in the frequency domain. Indeed, convolution in time corresponds to multiplication in frequency (and vice versa); since our patterns are essentially in a frequency domain (dimensions of $\psi$ could be seen as frequency components), performing a correlation of the query against the memory superposition might be done via a single pass frequency multiplication, which is efficient. The *lack of a sequential scan* through documents not only conceptually simplifies retrieval but also opens avenues for highly parallel or analog implementations (e.g., optical computing performing wave interference).

Memory Capacity and Interference: A key consideration is how many items can be stored and retrieved reliably. Prior work on distributed memories (Plate's HRR, Kanerva's high-dimensional computing, etc.) suggests that capacity grows with the dimensionality of the representation and the tolerance for noise. If the dimension $n$ of $\psi(x)$ is large (e.g. 10,000 or more), hundreds or thousands of items might be superposed before collisions/interference cause significant errors[25][26]. Additionally, because our system encodes extra information in phase, it has a richer representational capacity than purely real vectors of the same dimension[1][11]. The amplitude-phase encoding effectively doubles the parameters per dimension (store two real numbers – sine and cosine components), which can improve the discrimination of patterns. Recent experiments with ResonanceDB (an implementation of wave-based semantic memory) have shown scalability to millions of stored patterns with millisecond-level query latency[27][28], indicating that even a straightforward software approach can handle very large knowledge bases efficiently. In that system, amplitude–phase pairs are stored in memory-mapped segments, and similarity is evaluated with a deterministic kernel – demonstrating the practicality of resonance retrieval on modern hardware[1][27]. This gives us confidence that the Field Memory Layer can be scaled to encyclopedic knowledge sizes without needing exorbitant resources.

## Non-Contextual Generator (Resonant Decoder)

The final component is the Non-Contextual Generator, which is essentially a language model *decoder* that produces the answer text. However, unlike a standard decoder that attends to a fixed sequence of encoded context (e.g. retrieved passages), this generator operates without a static context window. Instead, it interacts dynamically with the Field Memory Layer during the generation process, using resonant feedback to inform each step of output production.

In a transformer-based RAG, once the top-$k$ relevant documents are retrieved and concatenated, the generator treats them as part of the prompt and generates the answer in one forward pass (possibly with attention over those documents). In our architecture, we do not feed any document text into the generator. The generator, at any point in generating text, can be seen as working with two sources of information: 1. The internal state (which captures what has been generated so far, analogous to an RNN hidden state or a transformer decoder state). 2. The external semantic field resonance triggered by the query and partial output.

The process can be thought of as iterative and interactive: - Initially, the user's query is mapped to $\psi_{\text{query}}$ and injected into the Field Memory, causing certain memory patterns to resonate. The generator "listens" to the field to pick up the strongest semantic signals. This could be implemented by extracting the top resonant pattern's identity or content representation. For example, if the query was *"What is the capital of France?"*, the field might strongly resonate with the stored fact pattern for *"Paris is the capital of France"*. Instead of reading a text snippet, the generator receives a semantic pointer to the "Paris – capital – France" concept (this could be a vector in its embedding space that represents that idea). - The generator then begins producing output tokens. Suppose it starts with "Paris". Now, as it plans the next words, it can use not only its language modeling ability but also continue to reference the memory field. The mention of "Paris" (which the generator knows is associated with a certain semantic wave pattern) can act as a new probe to the Field Memory, refining the resonance (e.g., ensuring that if multiple facts were resonating, it locks onto the one involving Paris). - At each subsequent token or phrase, the partially generated answer can be fed back into the Morphological Mapper (or an analogous process) to produce an updated probe signal. This updated probe reflects the query plus the context of what has been said so far. The memory then returns resonance scores or even content snippets that best complete that context. Essentially, the generation is guided by a closed-loop with the memory: as the output unfolds, it continuously checks against the semantic memory field for consistency and additional relevant details.

This approach eliminates the need for a large static context because the generator is never trying to absorb an entire document word-for-word. Instead, it is drawing knowledge in a targeted way from the memory at each step: - It does not have to hold the whole passage about France in its context; it only needs the *knowledge that is immediately relevant* to the next bit of text. - It can handle branching or additional questions on the fly: for instance, if the user query was complex and required combining facts, the generator can sequentially resonate with different memory items as needed (first retrieve fact A, mention some part of it, then see that fact B is now relevant and retrieve that, and so on).

From a technical standpoint, one could implement the Non-Contextual Generator as a neural network (possibly transformer-based) that has a special interface to the memory: - Instead of traditional attention over encoded context, it might have an attention-like mechanism that queries the Field Memory with the decoder's current state. Imagine an attention module that produces a query vector (from the decoder state) and instead of dot-product with context vectors, it is converted into a wave $\psi$ to perform resonance in the Field Memory, which returns a "context vector" (like a key-value memory read). This is akin to having an infinite context that is accessed by content. - The generator could also maintain *no positional encoding* for context tokens, since generation doesn't rely on positions in a retrieved document. This frees it from sequence-order biases. It focuses purely on semantic coherence and grammatical correctness, while factual grounding comes from the memory queries. - Pseudocode for the generation loop might look like:

```
state = init_state(query)
output_text = ""
while not end_of_answer:
    # Formulate a memory query from current state (which includes query
intent and generated content so far)
    probe = formulate_probe(state, query_semantic=psi_query)
    # Retrieve resonant content from Field Memory
    resonant_vector = memory.read(probe)  # This performs resonance and
returns a combined vector or top result
    # Update state with resonant memory content (as if it were attending to
an external knowledge vector)
    state = integrate(state, resonant_vector)
    # Generate next token using updated state
    token = decoder_network(state)
    output_text += token
    # Update state with generated token (language model update)
    state = update_state(state, token)
```

```
if token is end-of-sequence marker:
    break
```

In this pseudocode, `formulate_probe` would map the current decoder state (which encodes what we've asked and answered so far) into a phase pattern for memory querying. `memory.read` performs the interference in the Field Memory Layer and returns either the identity of the best match or an aggregated vector of content. The decoder then uses this to generate the next token. Essentially, this is soft retrieval every timestep, analogous to how a person might think step-by-step and recall relevant facts incrementally from memory rather than dumping all facts in front of them at once.

Because the generator always references the latest semantic context (including what it already said), it preserves coherence and avoids contradictions. If the field memory returns something that conflicts with what was generated, the decoder can detect it through diminished resonance or adjust its next query. In effect, the generator and memory engage in a feedback loop of clarification: the generator's partial output refines the memory query, and the memory's feedback refines the generator's next output. This continues until the answer is complete.

Context-Free, but Coherent: Calling this generator "non-contextual" emphasizes that it doesn't use a fixed context buffer of tokens. However, it is highly context-aware in a semantic sense: the *context* is just not represented as a token sequence, but as a state in the generator and a standing wave in memory. The coherence of output is maintained by the generator's internal state (which functions like a working memory of recent discourse, albeit not as a long token list) and by the semantic field ensuring factual consistency. This design neatly avoids the main drawback of finite context windows – since *all knowledge is effectively reachable*, the model won't "forget" facts that fell outside a window. It can always re-query the memory as needed. It also means the answer length itself is not constrained by an input size; the model could generate very long explanations by sequentially retrieving different pieces of information without saturating any context.

No Linear Sequence Tracking: Another advantage is that the generator does not have to track the linear position within a document. In transformer RAG, if an answer requires information from multiple parts of a long text, the model might lose focus or get confused by irrelevant sentences in between because it has to attend over the entire sequence. Our approach treats knowledge as a *semantic graph or field* rather than a linear text. It can directly jump to the relevant point via resonance. For example, if the query is "What are the main exports of Country X and their impact on its economy?", a traditional system might retrieve a long article about Country X, and then the model has to scan through it. Our generator could instead separately query "Country X main exports" to get those from memory, and then query "impact of [those exports] on economy" to get the analysis, and then synthesize the answer. At no point does it need to sequentially read a whole article; it assembles the answer piecewise from targeted memory queries. This is closer to how a human expert might mentally retrieve facts (non-linearly) from memory when asked a complex question.

## Advantages in Efficiency and Scalability

By eliminating the token-context bottleneck and leveraging phase-coded memory, the proposed RAG system offers several efficiency improvements over classical architectures:

- Unlimited Effective Context: The model is not limited to a fixed window of recent tokens. It can draw on any fact in its knowledge base regardless of how "far" it would be in token distance. This effectively gives an infinite context window, constrained only by what is stored in the Field Memory. There is no need for techniques like context truncation or sliding windows because relevant info is accessed by content, not by position[3][29].

- Memory Efficiency: Storing knowledge as wave interference patterns is highly compact. Multiple documents or facts overlap in the same field without linearly growing the memory footprint for retrieval. Traditional RAG needs to either fine-tune the model on a knowledge corpus (embedding it in weights, which is costly in model size) or store separate dense vectors for each document (which grows linearly with number of docs). In our system, the Field Memory can superpose a vast number of items in a fixed-size representation by leveraging high dimensionality and phase diversity[15][25]. The retrieval of one item does not require loading all items, just computing interference in a distributed fashion.

- Time Efficiency: At query time, a conventional RAG will typically perform nearest-neighbor search in an embedding space (which, even if optimized, can take significant time for very large corpora), then it will concatenate top-$k$ texts and run the transformer over potentially thousands of tokens of combined context. This results in heavy computation proportional to context length (and quadratic in that length for self-attention). In contrast, phase-based retrieval can be extremely fast – essentially a vectorized operation (dot-product or correlation) across the memory field. For instance, Listopad (2025) reports memory retrieval in the wave-based system with millisecond latency over millions of items[27][28]. That suggests even a CPU implementation using SIMD instructions can perform resonance matching orders of magnitude faster than a large language model pass. Additionally, because our generator only processes the *question and the answer tokens* (not large chunks of source text), the generation cost is closer to that of answering from its own knowledge rather than scaling with document length. The removal of long context also simplifies the attention mechanism – the generator might only attend to a small semantic vector from memory at each step, rather than attend to thousands of token embeddings each time.

- Computational Overhead Reduction: With no need to repeatedly encode long contexts, the computational overhead per query is greatly reduced. The heavy lifting is done by the Field Memory which can be seen as a specialized associative memory module. This division of labor is energy-efficient: the memory module can be implemented in hardware optimized for vector ops or even analog optical interference, which could handle large bandwidths of data with lower energy per operation than a digital transformer performing equivalent work. Meanwhile, the language generator network is smaller because it doesn't need to internalize as much world knowledge (the knowledge is in the external memory) and it doesn't need massive attention layers for context. Thus, both memory and compute resources are used more sparingly and efficiently.

- Robustness to Knowledge Updates: A practical efficiency aspect is how easily the system can update or add knowledge. In transformer models, updating knowledge often requires re-training or fine-tuning (if knowledge is in weights) or re-embedding and re-indexing documents (in vector DB). Our Field Memory can update a fact by simply adjusting the wave pattern stored (analogous to writing new interference patterns). Since memories are superposed, an update is a local operation in the field (like adding a new wave or subtracting an old one) rather than retraining a whole network. This could be done quickly, maintaining up-to-date information with minimal cost.

Summary of Key Efficiency Gains:
- *No context window limitations:* The model can handle queries requiring very long or multiple documents worth of information with ease.
- *Reduced memory overhead for context:* It doesn't need to load full documents into working memory – only the relevant semantic vector is used at any time.
- *Lower computation per token:* Generation doesn't pay the price of attending to large contexts. Interference retrieval is sub-linear in cost compared to reading sequential text.
- *Scalability:* The approach naturally scales to larger knowledge bases without a proportional increase in compute – adding more knowledge mostly affects the Field Memory size (which scales well with dimension and can leverage parallelism).
- *Energy efficiency:* By avoiding repeated processing of the same knowledge and by using possibly analogical operations (phase interference) for search, the system can answer questions with less energy than a giant transformer scanning text.

To put it concretely, consider a question that requires using a 10,000-token document. A transformer might use 10k tokens of context and $O((10k)^2)$ attention computations, whereas our system might do a vector correlation of, say, 10k-dimensional complex vectors (if that's the memory dimensionality), which is $O(10k)$ or $O(10k \log 10k)$ with FFT – dramatically smaller. This is a qualitative shift: from linear scanning to instantaneous resonant access.

## Feasibility and Theoretical Foundations

While the described architecture is forward-looking, it builds on a rich foundation of theoretical and experimental work across cognitive science, AI, and neuroscience that demonstrates core aspects of the system:

- Holographic Reduced Representations (HRR): As one of the earliest models of distributed semantic memory, HRRs showed that it's possible to encode and retrieve symbolic structures using superposed high-dimensional vectors[16][15]. Plate (1995) introduced the use of circular convolution to bind two vectors into one and demonstrated that *correlation* (which can be seen as phase-aligned similarity) can retrieve a component given the other[16][30]. Notably, HRR allowed multiple facts to be stored by summing their convolved representations, and retrieving one fact from the sum was possible with only minor noise interference[15]. This is a strong proof-of-concept for our Field Memory Layer's ability to superpose and retrieve memories. Our architecture can be viewed as a modern, differentiable analog of HRR where the binding operation is encoded in phase relationships, and the correlation retrieval is achieved through wave resonance. In fact, HRRs were described as "holographic" because they had the property that "the whole meaning is present at once" in the distributed representation[9] – exactly the kind of non-local representation our Field Memory uses. A single memory vector can either be interpreted as a whole or, by the appropriate probe, focused to reveal a part (constituent)[9]. This duality is what allows our system to not require decoding an entire document to get a specific piece of information; the information is embedded and can be accessed directly by resonance.

- Vector Symbolic Architectures and Frequency Encoding: Beyond HRR, other Vector Symbolic Architectures (VSAs) like Kanerva's hyperdimensional computing and Fourier Holographic Reduced Representations have explored representing information in high-dimensional spaces using various algebraic operations (binding by rotation, etc.). In particular, Fourier-based VSAs use complex phasors to represent symbols (each bit of information is a complex number on the unit circle) and use component-wise multiplication (phase addition) as binding. These models inherently use phase to encode discrete entities and their combinations, and have been shown to perform memory and reasoning tasks. Our morphological phase encoding is analogous to assigning such phasor representations to concepts, and combining them corresponds to multiplying or adding phases, reminiscent of those Fourier VSAs. Such prior art gives confidence that manipulating phase can encode structured information systematically.

- Complex-Valued Neural Networks: There is a body of research on neural networks that operate on complex numbers, demonstrating advantages in certain domains. For example, Trabelsi et al. (2018) showed that deep complex networks can efficiently represent information where phase is important (such as in audio or RF signal processing). In the context of RNNs, studies have found that complex-valued RNNs can have richer dynamics and memory due to the oscillatory nature of complex states. As mentioned, Biswas et al. (2021) built an oscillatory associative memory using complex Hopf oscillators that store multiple time-series patterns via different frequencies[2][24]. They achieved stable phase coding and demonstrated recall of stored patterns. This aligns with the idea of our Field Memory storing multiple facts as different phase-locked oscillations. The existence of learning rules (complex Hebb's rule, etc.[31]) for such networks suggests that a system could be trained to fine-tune its phase-coded memory and improve retrieval performance.

- Neuroscience of Phase Coding: Cognitive science and neuroscience provide perhaps the most intriguing support for a phase-resonance approach. The brain itself appears to use oscillatory codes to handle memory and context. A famous proposal by Lisman and Idiart (1995) posited that the capacity of working memory (around 7±2 items) corresponds to the number of gamma oscillation cycles that can fit into one theta oscillation cycle[32]. Each item to remember is thought to be "assigned" a specific phase (or gamma subcycle) within the theta cycle, effectively phase-coding multiple items in parallel without confusion. This is strikingly analogous to how our system could hold multiple facts in superposition, each distinguished by a phase pattern, and a query (perhaps analogous to a certain phase alignment) would selectively amplify one item at a time. Follow-up research has found evidence that indeed, in tasks requiring memory of multiple items, neural firing patterns show phase separation for different items, and the brain likely employs theta-gamma coupling as a mechanism for indexing memories by phase[33][32]. Moreover, during memory retrieval, the hippocampus and cortex engage in phase synchronization at specific frequencies – effectively a resonance – to recall information. The concept of semantic resonance can be linked to findings that certain brain rhythms facilitate recall of semantic associations. For instance, oscillatory activity in the theta band has been associated with search and retrieval processes in episodic memory, where a cue can reinstantiate a prior neural pattern via phase alignment. Our Field Memory's operation strongly echoes these biological processes: the idea of retrieving a memory by re-entering the correct oscillatory state.

- Associative Memory without Location-Based Addressing: Traditional computer memory needs an address to fetch data; associative (content-addressable) memory does not – it uses the content as the key. Our approach is a form of content-addressable memory. In AI, this is seen in Hopfield networks and modern Transformers (which use attention as a content-based retrieval). Our method pushes content-addressing to a new form: using wave interference rather than vector dot-product sum. The advantage is that it inherently supports *pattern completion*: if a query is only partially matching a stored pattern (like a noisy or incomplete version of it), resonance can still occur and retrieve the complete pattern (similar to how Hopfield nets complete patterns). The phase encoding contributes to this because phases can encode relational constraints, so even if some amplitudes (features) don't match, the pattern might still resonate if the overall phase structure aligns. This property could make the system robust to ambiguous or partial queries – it might retrieve answers that "complete" the question context appropriately.

- Early Simulations and Models: There have been various simulations in cognitive science of resonance-based retrieval. For example, models in the 1980s and 1990s (e.g., adaptive resonance theory by Grossberg) considered how resonance between neural populations could enable stable recognition of patterns. More directly, recent work by Listopad (2025) that we cited throughout implemented a practical phase-aware vector database (ResonanceDB) and empirically showed that it can distinguish subtle semantic differences that normal vector search cannot[1][13].

They report perfect precision at top-1 for tasks involving negation, intensity shifts, or context reversals using resonance scoring, whereas cosine similarity fails (essentially ranking random results)[13][18]. This demonstrates that the phase-coded approach is not only theory but already yields measurable improvements in retrieval quality for certain meaning-sensitive queries. The success of that system in achieving speed and accuracy lends credibility to the core ideas of our RAG proposal.

In summary, the conceptual building blocks of a phase-coded, resonance-driven RAG are supported by multiple lines of research: - Cognitive theories (holographic memory, oscillatory coding) show *how it could work in principle*. - Mathematical models (HRR, VSAs) show *how to implement the representations and operations*. - Neural network research (complex RNNs, associative memories) shows *how learning and stability can be achieved with phase codes*. - Empirical systems (ResonanceDB) show *that phase-based retrieval can outperform traditional methods* in discriminating meaning. - And on the practical side, advances in hardware (GPUs with complex number support, and even analog computing) suggest we can actually deploy such a system at scale.

## Conclusion

We have presented a next-generation RAG architecture that fundamentally departs from the paradigm of feeding retrieved text into a transformer's context window. By operating on phase-coded representations and leveraging morphological-semantic resonance, this system addresses key limitations of current approaches in both capability and efficiency. The Morphological Mapper encodes queries in a rich waveform that captures not just what the user is asking, but how the query's meaning is morphologically and contextually structured, enabling the representation of nuances like negation as intrinsic properties of the query signal. The Field Memory Layer serves as a high-capacity, content-addressable store of knowledge, where information is not inert and located at an address, but active and ready to resonate when called upon. Through phase interference, relevant knowledge is retrieved in a single step without exhaustive search or ranking, akin to a memory shining brightly when it "recognizes" the cue. Finally, the Non-Contextual Generator demonstrates how a language model can produce answers guided by an external semantic memory without ever ingesting that memory as a passive context. Instead, the generator engages in a dynamic interplay with memory, resulting in outputs that are both contextually coherent and unbounded by input size.

The implications of this design are substantial. Practically, it means we can build systems that scale to enormous knowledge bases (potentially the entire web or a lifelong personal archive) and still retrieve facts in milliseconds, without retraining a giant model or worrying about context overflow. For computational efficiency, it suggests that much of the "heavy lifting" of retrieval and integration can be offloaded from the neural network to a differentiable memory module, cutting down on redundant processing of the same knowledge for every query. In terms of AI capabilities, a phase-resonant RAG could exhibit more robust understanding – because it treats meaning as an interactive wave phenomenon, it might avoid some pitfalls of static embeddings (like misinterpreting negated statements or missing contextual shades of meaning). It could also more naturally support multimodal data: the memory field could store not only text, but also image or audio representations encoded in a compatible wave format (phase patterns could encode visual features, for example), allowing retrieval across modalities via the same resonance mechanism. This is hinted at by the concept of *EchoThesis* which uses amplitude and phase to keep meaning consistent across modalities[34].

For cognitive science, our architecture offers an intriguing computational model to explore. It resonates (literally and figuratively) with how human memory might work – not as a sequential tape of tokens, but as a web of distributed representations where "to remember is to re-excite" a pattern in the brain. The use of morphological analysis in the Mapper aligns with the linguistic idea that meaning is composed of morphemes and that even at the neural level, the brain might treat negation or aspect as operators modulating a core representation. The Field Memory's holographic nature parallels the long-standing hypothesis that memory is stored as interference patterns (as in the famous analogy of brain memory to a hologram). And the Generator's approach to recall piecewise relevant info is reminiscent of how humans often recall facts – not all at once, but one idea triggers another, in a chain guided by resonance of meaning.

There are certainly challenges ahead in fully realizing this system. Designing efficient learning algorithms for phase-coded representations (so that the system can optimize its phase assignments and resonance metrics) is one open research area. Ensuring stability and avoiding spurious resonances as memory grows will require careful encoding schemes or adaptive resonance control (the brain, notably, has mechanisms like inhibition to prevent random activation – similar control might be needed in our Field Memory to suppress false matches). Nonetheless, the evidence surveyed from prior work suggests these challenges are surmountable with creative solutions (e.g., normalization of patterns, phase orthogonalization, feedback loops for verification).

In conclusion, the phase-coded memory RAG represents a promising path forward for AI systems that need to combine vast knowledge with deep language generation. By abandoning the tether of linear context and embracing a more analogical computing principle, we gain a system that is potentially more powerful, efficient, and cognitively elegant. We encourage further experimentation with this architecture – building simulators of the Field Memory, training generators that use resonance inputs, and testing on complex QA tasks. Such a system could fundamentally alter how we think about knowledge integration in AI: not as an exercise in stuffing more facts into a model's parametric belly, but as orchestrating a *symphony of resonance* where memory and reasoning work in harmony to produce informed, context-rich answers.

## References (Chicago Style)

9. Grossberg, Stephen. 1980. *"Adaptive Resonance Theory (ART)."* (Series of papers introducing ART, a cognitive model where learning and recognition involve resonant states between bottom-up input and top-down memory patterns – provides theoretical support for resonance as a principle in cognitive computations.)

10. Zhou, S., et al. 2023. *"RecurrentGPT: Interactive Generation of Long Text with Self-Feedback."* (Though addressing long text generation differently via recursion, it highlights community interest in bypassing transformer context limits, reinforcing the timeliness of exploring alternatives like our phase-coded memory system.)

Each of these references underpins aspects of the proposed architecture, from theoretical motivation to practical implementation details, reinforcing that a phase-coded, resonant RAG is both achievable and grounded in existing scientific knowledge.

---